\documentclass[11pt,letterpaper]{article}
\usepackage[letterpaper]{geometry}
\usepackage{acl2012}
\usepackage{times}
\usepackage{latexsym}
\usepackage{amsmath}
\usepackage{multirow}
\usepackage{url}
\usepackage{subcaption}
\usepackage{float}

\usepackage{tikz}
\usepackage{CJK}
\usepackage{amssymb}

\usetikzlibrary{fit,positioning,arrows.meta}
\usepackage{float}
\usepackage{tabularx}
\usepackage{graphicx}

\usepackage{siunitx}
\usepackage{dcolumn}
\usepackage{pgfplots}
\usetikzlibrary{plotmarks}
\pgfplotsset{compat=1.7}
\newcolumntype{d}[1]{D{.}{.}{#1}}

\usepackage{color}
\usepackage{xcolor}
\usepackage{pdfpages}
\usepackage{colortbl}

\usepackage{arabtex}
\usepackage{utf8}
\usepackage{setspace}

\usepackage{changepage, lipsum}

\tikzset{neuron/.style={shape=circle, minimum size=1.25cm, 
  inner sep=0, draw, font=\small}, io/.style={neuron, fill=gray!20}}

\makeatletter
\newcommand{\@BIBLABEL}{\@emptybiblabel}
\newcommand{\@emptybiblabel}[1]{}

\makeatother
\usepackage[hidelinks]{hyperref}

\usepackage[inline]{enumitem}

\makeatletter
\newcommand*{\textoverline}[1]{$\overline{\hbox{#1}}\m@th$}
\makeatother

\title{Universal Word Segmentation: Implementation and Interpretation}

\author{Yan Shao, Christian Hardmeier, Joakim Nivre \\
Department of Linguistics and Philology, Uppsala University \\
 {\tt \{yan.shao, christian.hardmeier, joakim.nivre\}@lingfil.uu.se} \\
}
\date{}

\begin{document}
\maketitle
\begin{abstractX}
Word segmentation is a low-level NLP task that is non-trivial for a considerable number of languages. In this paper, we present a sequence tagging framework and apply it to word segmentation for a wide range of languages with different writing systems and typological characteristics. Additionally, we investigate the correlations between various typological factors and word segmentation accuracy. The experimental results indicate that segmentation accuracy is positively related to word boundary markers and negatively to the number of unique non-segmental terms. Based on the analysis, we design a small set of language-specific settings and extensively evaluate the segmentation system on the Universal Dependencies datasets. Our model obtains state-of-the-art accuracies on all the UD languages.  It performs substantially better on languages that are non-trivial to segment, such as Chinese, Japanese, Arabic and Hebrew, when compared to previous work.
\end{abstractX}

\section{Introduction}

Word segmentation is the initial step for most higher level natural language processing tasks, such as part-of-speech tagging (POS), parsing and machine translation. It can be regarded as the problem of correctly identifying word forms from a character string.

Word segmentation can be very challenging, especially for languages without explicit word boundary delimiters, such as Chinese, Japanese and Vietnamese. Even for space-delimited languages like English or Russian, relying on white space alone generally does not result in adequate segmentation as at least punctuation should usually be separated from the attached words. For some languages, the space-delimited units in the surface form are too coarse-grained and therefore often further analysed, as in the cases of Arabic and Hebrew. Even though language-specific word segmentation systems are near-perfect for some languages, it is still useful to have a single system that performs reasonably with no or minimum language-specific adaptations. 

%Word is a single distinct meaningful element in natural languages. 
Word segmentation standards vary substantially with different definitions of the concept of a word. In this paper, we will follow the teminologies of Universal Dependencies (UD), where words are defined as basic syntactic units that do not always coincide with phonological or orthographic words. Some orthographic tokens, known in UD as multiword tokens, therefore need to be broken into smaller units that cannot always be obtained by splitting the input character sequence.\footnote{Note that this notion of multiword token has nothing to do with the notion of multiword expression (MWE) as discussed, for example, in \newcite{sag2002multiword}.}

%In Universal Dependencies (UD) \cite{nivre2016universal}, words are defined as basic syntactic units. The syntactic words do not always coincide with phonological or orthographic words as they are considered as the basic units between which dependency relations hold.  Some orthographic tokens, known as multiword tokens, are therefore processed into more fine-grained units for syntactic analysis in the following steps. They are either simply further segmented by splitting off clitics and contractions, for instance \textit{cannot} to \textit{can} and \textit{not} in English, or transduced into different forms, like \textit{des} to \textit{de} and \textit{les} in French. The segmented word forms in UD still keep morphological properties and processing multiword tokens in both segmental and non-segmental cases does not involve morphological analysis. In addition, multiple orthographic tokens are occasionally combined into single words, such as numerical expressions like \textit{20 000}, in which the white space becomes a part of the syntactic word. The multi-token words are defined in a narrow sense and different from general multiword expressions. 

To perform word segmentation in the UD framework, neither rule-based tokenisers that rely on white space nor the naive character-level sequence tagging model proposed previously \cite{xue2003chinese} are ideal. In this paper, we present an enriched sequence labelling model for universal word segmentation. It is capable of segmenting languages in very diverse written forms. Furthermore, it simultaneously identifies the multiword tokens defined by the UD framework that cannot be resolved simply by splitting the input character sequence. 
%The identified non-segmental multiword tokens are processed either by querying a dictionary or using an attention-based character encoder-decoder. The proposed sequence tagging approach is applicable to all languages represented in UD.
% without any language-specific adaptation. 
%Retaining the simplicity of regular sequence tagging models, 
We adapt a regular sequence tagging model, namely the bidirectional recurrent neural networks with conditional random fields (CRF) \cite{lafferty2001crf} interface as the fundamental framework (BiRNN-CRF) \cite{huang2015bidirectional} for word segmentation. 

%Additionally, we investigate the correlations between some typological factors and segmentation accuracy to explore and interpret the accuracy gaps between different languages, which is helpful  Our model achieves substantially better results than previous work \cite{udpipe:2017} on languages that are challenging for word segmentation with very few and simple language settings based on the analysis. 

The main contributions of this work include: \begin{enumerate}[noitemsep,topsep=0.1cm]
\item We propose a sequence tagging model for word segmentation, both for general purposes (mere splitting) and full UD processing (splitting plus occasional transduction). %The model is applicable to all languages without language-dependent settings. It performs substantially better with specific settings with respect to language groups.
\item We investigate the correlation between segmentation accuracy and properties of languages and writing systems, which is helpful in interpreting the gaps between segmentation accuracies across different languages as well as selecting language-specific settings for the model.
\item Our segmentation system achieves state-of-the-art accuracy on the UD datasets and improves on previous work \cite{udpipe:2017} especially for the most challenging  languages. 
\item We provide an open source implementation.\footnote{
https://github.com/yanshao9798/segmenter
%The URL is suppressed for anonymous review.
}
\end{enumerate}

\begin{figure*}
\centering
\includegraphics[height=7.5cm]{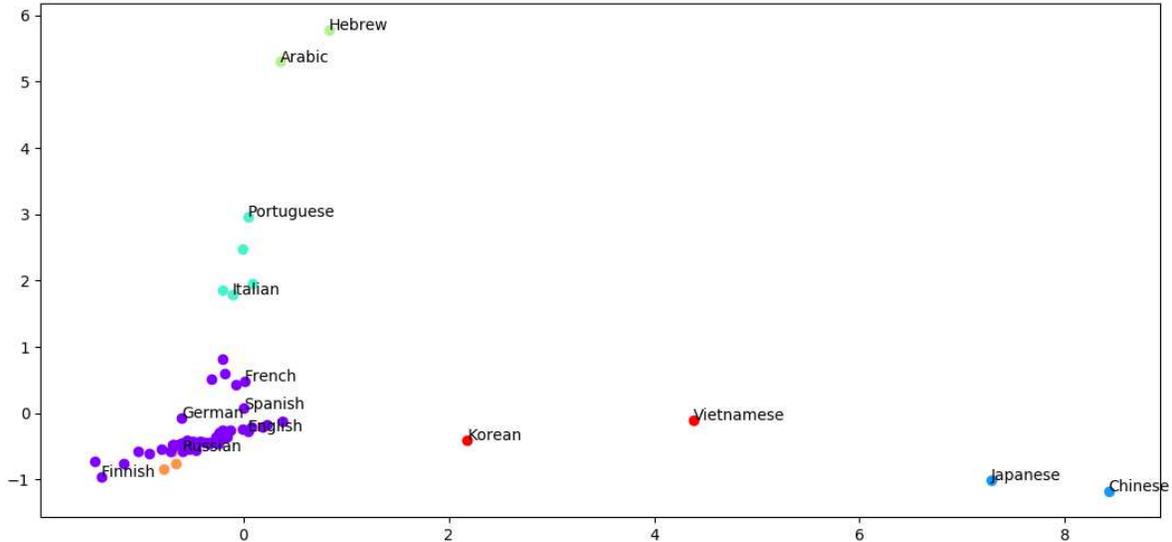}
\caption{K-Means clustering ($K=6$) of the UD languages. PCA is applied for dimensionality reduction.}\label{fig:3}
\end{figure*}

\section{Word Segmentation in UD}

The UD scheme for cross-linguistically consistent morphosyntactic annotation defines \emph{words} as syntactic units that have a unique part-of-speech tag and enter into syntactic relations with other words \cite{nivre2016universal}. For languages that use white\-space as boundary markers, there is often a mismatch between orthographic words, called \emph{tokens} in the UD terminology, and syntactic \emph{words}. Typical examples are clitics, like Spanish \emph{d{\'a}melo} = \emph{da me lo} (1 token, 3 words), and contractions, like French \emph{du} = \emph{de le} (1 token, 2 words). Tokens that need to split into multiple words are called \emph{multiword tokens} and can be further subdivided into those that can be handled by simple segmentation, like English \emph{cannot} = \emph{can not}, and those that require a more complex transduction, like French \emph{du} = \emph{de le}. We call the latter \emph{non-segmental multiword tokens}. In addition to multiword tokens, the UD scheme also allows  \emph{multitoken words}, that is, words consisting of multiple tokens, such as numerical expressions like \textit{20 000}.

%The syntactic words do not always coincide with phonological or orthographic words as they are considered as the basic units between which dependency relations hold.  Some orthographic tokens, known as multiword tokens, are therefore processed into more fine-grained units for syntactic analysis in the following steps. They are either simply further segmented by splitting off clitics and contractions, for instance \textit{cannot} to \textit{can} and \textit{not} in English, or transduced into different forms, like \textit{des} to \textit{de} and \textit{les} in French. The segmented word forms in UD still keep morphological properties and processing multiword tokens in both segmental and non-segmental cases does not involve morphological analysis. In addition, multiple orthographic tokens are occasionally combined into single words, such as numerical expressions like \textit{20 000}, in which the white space becomes a part of the syntactic word.

\section{Word Segmentation and Typological Factors} \label{sec:1}

%In this work, apart from investigating the impact of space delimiters on word segmentation, we analyse the correlation between segmentation accuracies on  different languages and several statistical factors that are related to linguistic typology and written system:
%To characterise the challenges of word segmentation posed by different languages, we propose several statistical factors with respect to linguistic typology and writing system:
We begin with the analysis of the difficulty of word segmentation. Word segmentation is fundamentally more difficult for languages like Chinese and Japanese because there are no explicit word boundary markers in the surface form \cite{xue2003chinese}. For Vietnamese, the space-segmented units are syllables that roughly correspond to Chinese characters rather than words. To characterise the challenges of word segmentation posed by different languages, we will examine several factors that vary depending on language and writing system. We will refer to these as \textbf{typological factors} although most of them are only indirectly related to the traditional notion of linguistic typology and depend more on writing system. 
\begin{itemize}[noitemsep,topsep=0.1cm]
\item \textbf{Character Set Size (CS)} is the number of unique characters, which is related to how informative the characters are to word segmentation. Each character contains relatively more information if the character set size is larger.
\item \textbf{Lexicon Size (LS)} is the number of unique word forms in a dataset, which indicates how many unique word forms have to be identified by the segmentation system. Lexicon size increases as the dataset grows in size.
\item \textbf{Average Word Length (AL)} is calculated by dividing the total character count by the word count. It is negatively correlated with the density of word boundaries. If the average word length is smaller, there are more word boundaries to be predicted. 
\item \textbf{Segmentation Frequency (SF)} denotes how likely it is that space-delimited units are further segmented. It is calculated by dividing the word count by the space-segment count.  
%as segments-space ratio and indicates how frequently the space-delimited units are further segmented. 
Languages like Chinese and Japanese have much higher segmentation frequencies than space-delimited languages.
\item \textbf{Multiword Token Portion (MP)} is the percentage of multiword tokens that are non-segmental.
\item \textbf{Multiword Token Set Size (MS)} is the number of unique non-segmental multiword tokens. 
\end{itemize}
The last two factors are specific to the UD scheme but can have a significant impact on word segmentation accuracy. 
%We assume that the UD-specific multiword tokens have a significant impact on word segmentation, which is expected to be characterised by MP and MS.

\begin{table}
\centering
\scalebox{0.75} {
\begin{tabular}{c|c|c|c|c|c}
\hline
CS & LS & AL & SF & MP & MS \\
\hline
0.058 & 0.938 & 0.101 & -0.043 & -0.060 & -0.028 \\
\hline
\end{tabular}
}
\caption{Pearson product-moment correlation coefficients between dataset size and the statistical factors.} \label{tab:a1}
\end{table}

%Detailed statistics of the proposed factors on the UD datasets are available in Table \ref{tab:7} in the Appendix. 
All the languages in the UD dataset are characterised and grouped by the typological factors %as shown 
in Figure \ref{fig:3}. We standardise the statistics $x$ of the proposed factors on the UD datasets with the arithmetic mean $\mu$ and the standard deviation $\sigma$ as $\frac{x-\mu}{\sigma}$. We use them as features and apply K-Means clustering ($K=6$) to group the languages.  Principal component analysis (PCA) \cite{abdi2010principal} is used for dimensionality reduction and visualisation. 

The majority of the languages in UD are space-delimited with few or no multiword tokens and they are grouped at the bottom left of Figure \ref{fig:3}. They are statistically similar from the perspective of word segmentation. The Semitic languages Arabic and Hebrew with rich non-segmental multiword tokens are positioned at the top. In addition, languages with large character sets and high segmentation frequencies, such as Chinese, Japanese and Vietnamese are clustered together. Korean is distanced from the other space-delimited languages as it contains white-space delimiters but has a comparatively large character set. Overall, the $x$-axis of Figure \ref{fig:3} is primarily related to character set size and segmentation frequency, while the $y$-axis is mostly associated with multiword tokens. 

\begin{table}
\centering
\scalebox{0.70} {
\begin{tabular}{c|r|r|r|r|r|r}
\hline
Language & CS & LS & AL & SF & MP & MS \\
%\hline
%Ancient Greek & 184 & 33,237 & 4.6 & 1.25 & 0.0 & 0 \\
%Ancient Greek-PROIEL & 175 & 29,963 & 5.06 & 1.09 & 0.0 & 0\\
\hline
Czech & 140 & 125,342 & 4.83 & 1.26 & 0.0018 & 9 \\
Czech-CAC & 93 & 66,256 & 5.06 & 1.20 & 0.0022 & 12\\
Czech-CLIT & 96 & 2,774 & 5.30 & 1.14 & 0.0005 & 1 \\
\hline
English & 108 & 19,672 & 4.06 & 1.24 & 0.0 & 0 \\
English-LinES &  82 & 7,436 & 4.01 & 1.22 & 0.0 & 0 \\
English-ParTUT & 94 & 5,532 & 4.50 & 1.22 & 0.0002 & 6 \\
\hline
Finnish & 244 & 49,210 & 6.49 & 1.28 & 0.0 & 0 \\
Finnish-FTB & 95 & 39,717 & 5.94 & 1.14 & 0.0 & 0 \\
\hline 
French & 298 & 42,250 & 4.33 & 1.27 & 0.0281 & 9 \\
French-ParTUT & 96 & 3,364 & 4.53 & 1.27 & 0.0344 & 4 \\
French-Sequota & 108 & 8,452 & 4.48 & 1.29 & 0.0277 & 7 \\
\hline 
Latin & 57 & 6,927 & 5.05 & 1.28 & 0.0 & 0 \\
Latin-ITTB & 42 & 12,526 & 5.06 & 1.24 & 0.0 & 0 \\
\hline
Portuguese & 114 & 26,653 & 4.15 & 1.32 & 0.0746 & 710 \\
Portuguese-BR & 186 & 29,906 & 4.11 & 1.29 & 0.0683 & 35 \\
\hline
Russian & 189 & 25,708 & 5.21 & 1.26 & 0.0 & 0 \\
Russian-SynTagRus & 157 & 107,890 & 5.12 & 1.30 & 0.0 & 0 \\
\hline
Slovenian & 99 & 29,390 & 4.63 & 1.23 & 0.0 & 0 \\
Slovenian-SST & 40 & 4,534 & 4.29 & 1.12 & 0.0 & 0 \\
\hline
Swedish & 86 & 12,911 & 4.98 & 1.20 & 0.0 & 0 \\
Swedish-LinES & 86 & 9,659 & 4.50 & 1.19 & 0.0 & 0 \\
\hline
\end{tabular}
}
\caption{Different UD datasets in same languages and the statistical factors.} \label{tab:a2}
\end{table}

\begin{figure*}[h]
    \centering
    \scalebox{0.78}{
    \fbox{
        \begin{tabular}{rl}
        Char. &\texttt{On consid\`{e}re qu'environ 50 000 Allemands du Wartheland ont p\'{e}ri pendant la p\'{e}riode.}\\
        Tags &\texttt{BEXBIIIIIIIEXBIEBIIIIIEXBIIIIEXBIIIIIIIEX\textoverline{BE}XBIIIIIIIIEXBIEXBIIEXBIIIIIEXBEXBIIIIIES}\\
        \end{tabular}
    }
    }
    \caption{Tags employed for word segmentation. \textit{50 000} is a multitoken word, while \textit{qu'environ} and \textit{du} are multiword tokens that should be processed differently.}
    \label{fig:1}
\end{figure*}

Dataset sizes for different languages in UD vary substantially. Table \ref{tab:a1} shows the correlation coefficients between the dataset size in sentence number and the six typological factors. Apart from the lexicon size, all the other factors, including multiword token set size, have no strong correlations with dataset size. From Table \ref{tab:a2}, we can see that the factors, except for lexicon size, are relatively stable across different UD treebanks for the same language, which indicates that they do capture properties of these languages, although some variation inevitably occurs due to corpus properties like genre. 
 %characterise the properties of different languages in general, but they may still be incidental and related to the genres of the datasets. 
%This shows that the majority of the proposed factors are typological, rather than dataset specific. 

In this paper, we thoroughly investigate the correlations between the proposed statistical factors and segmentation accuracy. Moreover, we aim to find specific settings that can be applied to improve segmentation accuracy for each language group.

\section{Sequence Tagging Model}\label{sec:4}

Word segmentation can be modelled as a character-level sequence labelling task \cite{xue2003chinese,chen2015long}. Characters as basic input units are passed  into a sequence labelling model and a sequence of tags that are associated with word boundaries are predicted. In this section, we introduce the boundary tags adopted in this paper.

Theoretically, binary classification is sufficient to indicate whether a character is the end of a word for segmentation. In practice, more fine-grained tagsets result in higher segmentation accuracy \cite{Y06-1012}. Following the work of \newcite{shao17}, we employ a baseline tagset consisting of four tags: {\tt B}, {\tt I}, {\tt E}, and {\tt S}, to indicate a character positioned at the beginning ({\tt B}), inside ({\tt I}), or at the end ({\tt E}) of a word, or occurring as a single-character word ({\tt S}). 

The baseline tagset can be applied to word segmentation of Chinese and Japanese without further modification.
 For languages with space-delimiters, we add an extra tag {\tt X} to mark the characters, mostly spaces, that do not belong to any words/tokens. As illustrated in Figure \ref{fig:1}, the regular spaces are marked with {\tt X} while the space in a multitoken word like \textit{50 000} is disambiguated with {\tt I}. 

\setcode{utf8}

To enable the model to simultaneously identify non-segmental multiword tokens for languages like Spanish and Arabic in the UD framework, we extend the tagset by adding four tags {\tt \textoverline{B}}, {\tt \textoverline{I}}, {\tt \textoverline{E}}, {\tt \textoverline{S}} that correspond to {\tt B}, {\tt I}, {\tt E}, {\tt S} to mark corresponding positions in non-segmental multiword tokens and to indicate their occurrences. As shown in Figure \ref{fig:1}, the multiword token \textit{qu'environ} is split into \textit{qu'} and \textit{environ} and therefore the corresponding tags are {\tt BIEBIIIIIE}. This contrasts with \textit{du}, which should be transduced into \textit{de} and \textit{le}. Moreover, the extra tags disambiguate whether the multiword tokens should be split or transduced according to the context. For instance, {\setnastaliq
{\scriptsize
 \<ومما> 
} 
 \textit{(wamimma)} in Arabic is occasionally split into 
 {\scriptsize
 \<و> 
} 
 \textit{(wa)} and 
 {\scriptsize
 \<مما> 
} 
 \textit{(mimma)} but more frequently transduced into 
 {\scriptsize
 \<و> 
}
 \textit{(wa)}, 
 {\scriptsize
 \<من> 
} 
 \textit{(min)} and 
 {\scriptsize
 \<ما> 
} 
 \textit{(ma)}
}. The corresponding tags are {\tt SBIE} and {\tt \textoverline{BIIE}}, respectively. The transduction of the identified multiword tokens will be described in detail in the following section. 

The complete tagset is summarised in Table \ref{tab:1}. The proposed sequence model can easily be extended to perform joint sentence segmentation by adding two more tags to mark the last character of a sentence \cite{uu-conll17}. {\tt T} is used if the character is a single-character word and {\tt U} otherwise. {\tt T} and {\tt U} can be used together with  {\tt B}, {\tt I}, {\tt E}, {\tt S}, {\tt X} for general segmentation, or with {\tt \textoverline{B}}, {\tt \textoverline{I}}, {\tt \textoverline{E}}, {\tt \textoverline{S}} additionally for full UD processing. Joint sentence segmentation is not addressed any further in this paper. 

\begin{table}
\centering
\scalebox{0.85}{
\begin{tabular}{l|c|c}
\hline
& Tags & Applied Languages \\
\hline
Baseline Tags & B, I, E, S & Chinese, Japanese, ...\\
Boundary & X & Russian, Hindi, ...\\
Transduction & \textoverline{B}, \textoverline{I}, \textoverline{E}, \textoverline{S} & Spanish, Arabic, ...\\
\hline
Joint Sent. Seg. & T, U & All languages\\
\hline
\end{tabular}
}
\caption{Tag set for universal word segmentation.}\label{tab:1}
\end{table}

\section{Neural Networks for Segmentation}

\subsection{Main network}\label{sec:main}
\begin{CJK}{UTF8}{gbsn}
\begin{figure}[t]
\begin{center}
\scalebox{0.74}{
\begin{tikzpicture}[x=1.5cm, y=1.5cm]

\node (char1) {夏};
\node (char2) [right=1cm of char1]{天};
\node (char3) [right=1cm of char2]{太};
\node (char4) [right=1cm of char3]{热};

\node (eng2) [above=0.2cm of char3]{(too)};
\node (eng3) [above=0.2cm of char4]{(hot)};
\node (eng1) [left=1cm of eng2]{(summer)};

\foreach \x in {1,...,4}
	\node [circle, draw, minimum size=0.8cm] (emb1-\x) [below=0.2cm of char\x]{};
\foreach \x in {1,...,4}	
	\node [circle, draw, minimum size=0.8cm] (emb2-\x) [below=0.05cm of emb1-\x]{};
\foreach \x in {1,...,4}	
	\node [circle, draw, minimum size=0.8cm] (emb3-\x) [below=0.05cm of emb2-\x]{};
\foreach \x in {1,...,4}	
	\node[fit=(emb1-\x)(emb3-\x),inner sep=0, draw](emb-\x) {};

\node (note1) [left=0.3cm of emb2-1, text width=2cm, align=center]{character \\ representations};

\foreach \x in {1,...,4}
	\node [circle, draw, minimum size=0.8cm] (fgru-\x) [below=1cm of emb-\x]{GRU};	

\coordinate  [draw=none, left=0.6cm of fgru-1](fgru-0) ;
\coordinate  [draw=none, right=0.6cm of fgru-4](fgru-5);

\foreach \x [count=\xi from 1] in {0,...,4}
	\draw [-{Latex[length=2mm]}] (fgru-\x) -- (fgru-\xi);
 
\foreach \x in {1,...,4}
	\draw [-{Latex[length=2mm]}] (emb3-\x) -- (fgru-\x)[dashed];

\node (note2) [left=0.3cm of fgru-1, text width=2cm, align=center]{forward\\ RNN};

\foreach \x in {1,...,4}
	\node [circle, draw, minimum size=0.8cm] (bgru-\x) [below=1cm of fgru-\x]{GRU};

\coordinate  [draw=none, left=0.6cm of bgru-1](bgru-0) ;
\coordinate  [draw=none, right=0.6cm of bgru-4](bgru-5);

\foreach \x [count=\xi from 1] in {0,...,4}
	\draw [{Latex[length=2mm]}-] (bgru-\x) -- (bgru-\xi);

\foreach \x in {1,...,4}
	\draw[-{Latex[length=2mm]}] (emb3-\x.south) to [out=210,in=135] (bgru-\x.north)[dashed] ;

\node (note3) [left=0.3cm of bgru-1, text width=2cm, align=center]{backward\\ RNN};

\node [circle, draw, minimum size=1.2cm] (crf-1) [below=1cm of bgru-1]{B};
\node [circle, draw, minimum size=1.2cm] (crf-2) [below=1cm of bgru-2]{E};
\node [circle, draw, minimum size=1.2cm] (crf-3) [below=1cm of bgru-3]{S};
\node [circle, draw, minimum size=1.2cm] (crf-4) [below=1cm of bgru-4]{S};

\node (note4) [left=0.3cm of crf-1, text width=2cm, align=center]{CRF\\ layer};

\foreach \x in {1,...,4}
	\draw [-{Latex[length=2mm]}] (bgru-\x) -- (crf-\x)[dashed];

\foreach \x in {1,...,4}
	\draw[-{Latex[length=2mm]}] (fgru-\x.south) to [out=335,in=40] (crf-\x.north)[dashed] ;

\foreach \x [count=\xi from 2] in {1,...,3}
	\draw [-] (crf-\x) -- (crf-\xi);

\node (post-3) [below=0.6cm of crf-3]{太};
\node (post-4) [below=0.6cm of crf-4]{热};
\node (post-12) [left=1cm of post-3]{夏天};

\node (note4) [left=1.7cm of post-12, text width=2cm, align=center]{output};

\end{tikzpicture}
}
\end{center}
\caption{The BiRNN-CRF model for segmentation. The dashed arrows indicate that dropout is applied.}\label{fig:2}
\end{figure}
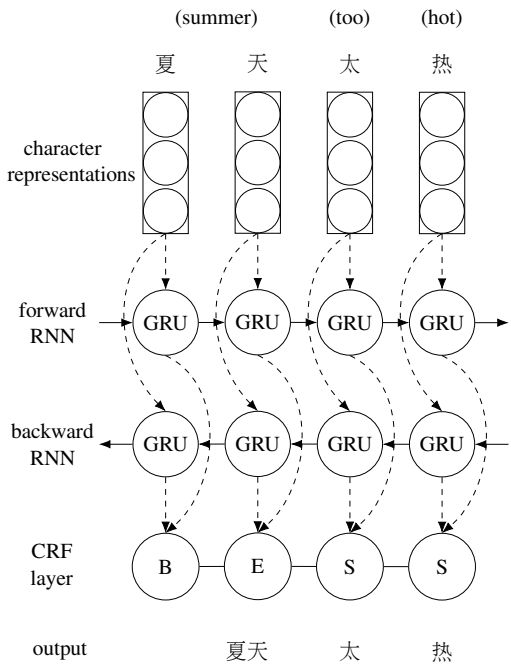

\begin{figure}[t]
\centering
\scalebox{0.75}{
\begin{tikzpicture}[x=1.5cm, y=1.5cm]

\node (char1) {夏};
\node (char2) [right=1.2cm of char1]{天};
\node (char3) [right=1.2cm of char2]{太};
\node (char4) [right=1.2cm of char3]{热};

\node (eng2) [above=0.4cm of char3]{(too)};
\node (eng3) [above=0.4cm of char4]{(hot)};
\node (eng1) [left=1.15cm of eng2]{(summer)};

\node[fit=(char3)(char3),inner sep=0, draw](char33)[dashed] {};
\node[fit=(char2)(char3),inner sep=0.1cm, draw](char23){};
\node[fit=(char2)(char4),inner sep=0.2cm, draw](char24)[dashed] {};

\node [circle, draw, minimum size=0.8cm] (gram1) [below=2cm of char2]{};
\node[fit=(gram1)(gram1),inner sep=0, draw](gram11){};

\node [circle, draw, minimum size=0.8cm] (gram2) [right=0.2cm of gram11]{};
\node [circle, draw, minimum size=0.8cm] (gram3) [right=0.05cm of gram2]{};
\node[fit=(gram2)(gram3),inner sep=0, draw](gram22){};

\node [circle, draw, minimum size=0.8cm] (gram4) [right=0.2cm of gram22]{};
\node [circle, draw, minimum size=0.8cm] (gram5) [right=0.05cm of gram4]{};
\node [circle, draw, minimum size=0.8cm] (gram6) [right=0.05cm of gram5]{};
\node[fit=(gram4)(gram6),inner sep=0, draw](gram33){};

\draw[-{Latex[length=2mm]}] (char33.south) to [out=210,in=95] (gram11.north);
\draw[-{Latex[length=2mm]}] (char23.south) to [out=190,in=95] (gram22.north);
\draw[-{Latex[length=2mm]}] (char24.south) to [out=270,in=95] (gram33.north);

\node (vec1) [below=0.4cm of gram11]{$V_{i, i}$};
\node (vec2) [below=0.4cm of gram22]{$V_{i-1, i}$};
\node (vec3) [below=0.4cm of gram33]{$V_{i-1, i+1}$};

\node (note1) [left=0.8cm of gram11, text width=2cm, align=center]{n-gram character\\ representation $V_3$};

\end{tikzpicture}
}
\caption{Concatenated 3-gram model. The third character is the pivot character in the given context.}\label{fig:2a}
\end{figure}
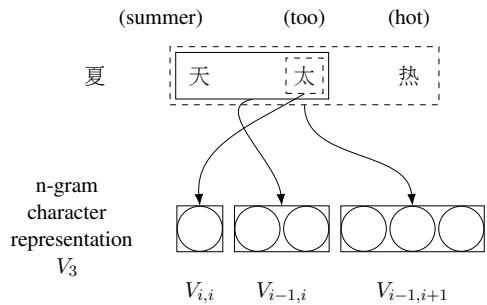

The main network for regular segmentation as well as non-segmental multiword token identification is an adaptation of the BiRNN-CRF model \cite{huang2015bidirectional} (see Figure \ref{fig:2}). 

The input characters can be represented as conventional character embeddings. Alternatively, we employ the concatenated 3-gram model introduced by \newcite{shao17}. In this representation (Figure \ref{fig:2a}), the pivot character in a given context is represented as the concatenation of the character vector representation along with the local bigram and trigram vectors. The concatenated n-grams encode rich local information as the same character has different yet closely related vector representations in different contexts. 
%We use single vectors to represent all the characters and n-grams that appear only once in the training set while training. 
For each n-gram order, we use a single vector to represent the terms that appear only once in the training set while training. These vectors are later used as the representations for unknown characters and n-grams in the development and test sets. All the embedding vectors are initialised randomly.

The character vectors are passed to the forward and backward recurrent layers. Gated recurrent units (GRU) \cite{cho2014properties} are employed as the basic recurrent cell to capture long term dependencies and sentence-level information. Dropout \cite{srivastava2014dropout} is applied to both the inputs and the outputs of the bidirectional recurrent layers. A first-order chain CRF layer is added on top of the recurrent layers to incorporate transition information between consecutive tags, which ensures that the optimal sequence of tags over the entire sentence is obtained. The optimal sequence can be computed efficiently via the Viterbi algorithm. % both for training and decoding. The time complexity is linear with respect to sentence length.

\subsection{Transduction}

The non-segmental multiword tokens identified by the main network are transduced into corresponding components in an additional step. Based on the statistics of the multiword tokens to be transduced on the entire UD training sets, 98.3\% only have one possible transduction, which indicates that the main ambiguity of non-segmental multiword tokens comes with identification, not transduction. We therefore transduce the identified non-segmental multiword tokens in a context-free fashion. For multiword tokens with two or more valid transductions, we only adopt the most frequent one.

In most languages that have multiword tokens, the number of unique non-segmental multiword tokens is rather limited, such as in Spanish, French and Italian. For these languages, we build dictionaries from the training data to look up the multiword tokens. However, in some languages like Arabic and Hebrew, multiword tokens are very productive and therefore cannot be well covered by dictionaries generated from training data. Some of the available external dictionary resources with larger coverage, for instance the MILA lexicon \cite{itai2008language}, do not follow the UD standards.

In this paper, we propose a generalising approach to processing non-segmental multiword tokens. If there are more than 200 unique multiword tokens in the training set for a language, we train an attention-based encoder-decoder \cite{bahdanau2014neural} equipped with shared long-short term memory cells (LSTM) \cite{hochreiter1997long}. At test time, identified non-segmental multiword tokens are first queried in the dictionary. If not found, the segmented components are generated with the encoder-decoder as character-level transduction. Overall, we utilise rich context to identify non-segmental multiword tokens, and then apply a combination of dictionary and sequence-to-sequence encoder-decoder to transduce them.

\subsection{Implementation}

\begin{table}
\centering
\scalebox{0.85}{
\begin{tabular}{l|c}
\hline
Character embedding size & 50 \\
\hline
GRU/LSTM state size & 200 \\
\hline
Optimiser & Adagrad \\
Initial learning rate (main) & 0.1 \\
Decay rate & 0.05 \\
Gradient Clipping & 5.0 \\
Initial learning rate (encoder-decoder) & 0.3 \\
\hline
Dropout rate & 0.5 \\
Batch size & 10 \\
\hline
\end{tabular}
}
\caption{Hyper-parameters for segmentation.}\label{tab:2}
\end{table}

Our universal word segmenter is implemented using the TensorFlow library \cite{abadi2016tensorflow}. Sentences with similar lengths are grouped into the same bucket and padded to the same length. We construct sub-computational graphs for each bucket so that sentences of different lengths are processed more efficiently. 

Table \ref{tab:2} shows the hyper-parameters adopted for the neural networks. We use one set of parameters for all the experiments as we aim for a simple universal model, although fine-tuning the hyper-parameters on individual languages might result in additional improvements. The encoder-decoder is trained prior to the main network. The weights of the neural networks, including the embeddings, are initialised using the scheme introduced in \newcite{glorot2010understanding}. The network is trained using back-propagation. All the random embeddings are fine-tuned during training by back-propagating gradients. Adagrad  \cite{duchi2011adaptive} with mini-batches is employed for optimization. The initial learning rate $ \eta_0$ is updated with a decay rate $\rho$. 

The encoder-decoder is trained with the unique non-segmental multiword tokens extracted from the training set. 5\% of the total instances are subtracted for validation. The model is trained for 50 epochs and the score of how many outputs exactly match the references is used for selecting the weights. 

For the main network, word-level F1-score is used to measure the performance of the model after each epoch on the development set. The network is trained for 30 epochs and the weight of the best epoch is selected. %The training time is proportional to the size of the training set. 

To increase efficiency and reduce memory demand both for training and decoding, we truncate sentences longer than 300 characters. At decoding time, the truncated sentences are reassembled at the recorded cut-off points in a post-processing step. % after being processed.

\section{Experiments}

\subsection{Datasets and Evaluation}

Datasets from Universal Dependencies 2.0 \cite{nivre2016universal} are used for all the word segmentation experiments.\footnote{We employ the version that was used in the CoNLL 2017 shared task on UD parsing.} 
In total, there are 81 datasets in 49 languages that vary substantially in size. The training sets are available in 45 languages. 
%Uyghur and Kazakh are not used for training due to their small sizes. For the rest,
We follow the standard splits of the datasets. If no development set is available, 10\% of the training set is subtracted. 
%The test sets are reserved for final evaluation.

We adopt word-level precision, recall and F1-score as the evaluation metrics. The candidate and the reference word sequences in our experiments may not share the same underlying characters due to the transduction of non-segmental multiword tokens. The alignment between the candidate words and the references becomes unclear and therefore it is difficult to compute the associated scores. To resolve this issue, we use the longest common subsequence algorithm to align the candidate and the reference words. The matched words are compared and the evaluation scores are computed accordingly:
\begin{align}
R &= \frac{|c \cap r|}{|r|} \\
P &= \frac{|c \cap r|}{|c|} \\
F &= 2 \cdot \frac{R \cdot P}{R + P}
\end{align}
where $c$ and $r$ denote the sequences of candidate words and reference words, and $|c|$, $|r|$ are their lengths. $|c \cap r|$ is the number of candidate words that are aligned to reference words by the longest common subsequence algorithm. The word-level evaluation metrics adopted in this paper are different from the boundary-based alternatives \cite{palmer1997chinese}.

We adapt the evaluation script from the CoNLL 2017 shared task \cite{zeman-EtAl:2017:K17-3} to calculate the scores. In the following experiments, we only report the F1-score.

In the following sections, we thoroughly investigate correlations between several language-specific characteristics and segmentation accuracy. All the experimental results in Section \ref{sec:52} are obtained on the development sets. The test sets are reserved for final evaluation, reported in Section \ref{sec:63}.

\subsection{Language-Specific Characteristics}\label{sec:52}

\subsubsection{Word-Internal Spaces}

For Vietnamese and other languages with similar historical backgrounds, such as Zhuang and Hmongic languages \cite{zhou1991}, the space-delimited syllables containing no punctuation are never segmented but joined into words with word-internal spaces instead. The space-delimited units can therefore be applied as the basic elements for tag prediction if we pre-split punctuation. Word segmentation for these languages thus becomes practically the same as for Chinese and Japanese. 

Table \ref{tab:2a} shows that a substantial improvement can be achieved if we use space-delimited syllables as the basic elements for word segmentation for Vietnamese. It also drastically increases both training and decoding speed as the sequence of tags to be predicted becomes much shorter. 

\begin{table}
\centering
\scalebox{0.85}{
\begin{tabular}{l|c|c}
\hline
Basic Unit & F1-score & Training Time (s) \\
\hline
Latin Character & 82.79 & 572 \\
\hline
Space-delimited Unit & 87.62 & 218 \\
\hline
\end{tabular}
}
\caption{Different segmentation units employed for word segmentation on Vietnamese. Concatenated 3-gram is not used.}\label{tab:2a}
\end{table}

%\subsubsection{Productive multiword Tokens}

%multiword tokens are very productive in Arabic and Hebrew as we see that more than 3,000 unique multiword tokens are found in the training sets. The transducer yields relatively higher scores on Hebrew while it is more challenging to transcribe Arabic. 
%A substantial amount of multiword tokens are transcribed into several duplicates of themselves in Arabic, which is difficult for the transducer to handle properly. 

\subsubsection{Character Representation} \label{sec:ng}

\begin{figure}[t]
\centering
\scalebox{0.85}{
\begin{tikzpicture}
\begin{axis}[
%x=12cm,
xtick={1, 2,...,10},
ytick={0, 0.1, 0.2, 0.3, 0.4, 0.5, 0.6, 0.7, 0.8, 0.9, 1.0},   
ylabel=F1-Score,
xlabel=$N/300$, 
xmin=1,
xmax=10,
ymin=0.77,ymax=1.02,
ymajorgrids=true,
 xmajorgrids=true,
 grid style=dashed,
legend columns=3, 
%legend pos=outer north east,
%legend style={font=\fontsize{7}{6}}
legend style={at={(0.5, -0.2)},anchor=north}
]

\addplot[smooth,color=orange, fill opacity=0.1,mark=o] file {n3/Arabic.dat};
\addplot[smooth,color=teal, fill opacity=0.1, mark=square] file {n3/Catalan.dat};
\addplot[smooth,color=blue,mark=otimes] file {n3/Chinese.dat};
\addplot[smooth,color=red,mark=oplus] file {n3/English.dat};
\addplot[smooth,color=red!50!green,mark=x] file {n3/Japanese.dat};
\addplot[smooth,color=violet, mark=triangle] file {n3/Spanish.dat};

\addplot[dashed,smooth,color=orange, fill opacity=0.1,mark=o,mark options={scale=1,solid}] file {n1/Arabic.dat};
\addplot[dashed,smooth,color=teal, fill opacity=0.1, mark=square,mark options={scale=1,solid}] file {n1/Catalan.dat};
\addplot[dashed,smooth,color=blue,mark=otimes,mark options={scale=1,solid}] file {n1/Chinese.dat};
\addplot[dashed,smooth,color=red,mark=oplus,mark options={scale=1,solid}] file {n1/English.dat};
\addplot[dashed,smooth,color=red!50!green,mark=x,mark options={scale=1,solid}] file {n1/Japanese.dat};
\addplot[dashed,smooth,color=violet, mark=triangle,mark options={scale=1,solid}] file {n1/Spanish.dat};

\addlegendentry{Arabic}
\addlegendentry{Catalan}
\addlegendentry{Chinese}
\addlegendentry{English}
\addlegendentry{Japanese}
\addlegendentry{Spanish}

\end{axis}
\end{tikzpicture}
}
\caption{Segmentation results with unigram character embeddings (dashed) and concatenated 3-gram vectors for character representations with different numbers of training instances $N$.}\label{fig:5a}
\end{figure}
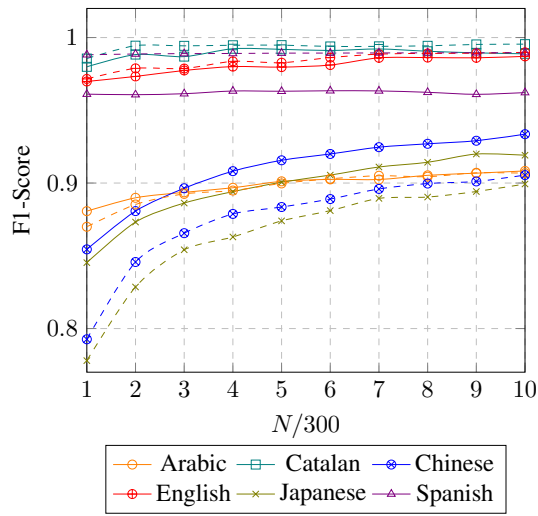

We apply regular character embeddings and concatenated 3-gram vectors introduced in Section \ref{sec:main} to the input characters and test their performances respectively. First, the experiments are extensively conducted on all the languages with the full training sets. The results show that the concatenated 3-gram model is substantially better than the regular character embeddings on Chinese, Japanese and Vietnamese, but notably worse on Spanish and Catalan. For all the other languages, the differences are marginal.

To gain more insights,  we select six languages, namely Arabic, Catalan, Chinese, Japanese, English and Spanish for more detailed analysis via learning curve experiments.  The training sets are gradually extended by 300 sentences at a time. The results are shown in Figure \ref{fig:5a}. Regardless of the amounts of training data and the other typological factors, concatenated 3-grams are better on Chinese and Japanese and worse on Spanish and Catalan. We expect the concatenated 3-gram representation to outperform simple character embeddings on all languages with a large character set but no space delimiters.

Since adopting the concatenated 3-gram model drastically enlarges the embedding space, in the following experiments, including the final testing phase, concatenated 3-grams are only applied to Chinese, Japanese and Vietnamese. 

\subsubsection{Space Delimiters}\label{sec:sd}

\begin{figure}[t]
\centering
\scalebox{0.85}{
\begin{tikzpicture}
\begin{axis}[
%x=12cm,
xtick={1, 2,...,10},
ytick={0, 0.1, 0.2, 0.3, 0.4, 0.5, 0.6, 0.7, 0.8, 0.9, 1.0},    
ylabel=F1-Score,
xlabel=$N/300$, 
xmin=1,
xmax=10,
ymin=0.55,ymax=1.02,
ymajorgrids=true,
 xmajorgrids=true,
 grid style=dashed,
legend columns=3, 
%legend pos=outer north east,
%legend style={font=\fontsize{7}{6}}
legend style={at={(0.5, -0.2)},anchor=north}
]

\addplot[smooth,color=orange, fill opacity=0.1,mark=o] file {ip/Arabic.dat};

\addplot[smooth,color=blue,mark=otimes] file {sp/Chinese.dat};
\addplot[smooth,color=red,mark=oplus] file {ip/English.dat};
\addplot[smooth,color=red!50!green,mark=x] file {ip/Korean.dat};
\addplot[smooth,color=teal, fill opacity=0.1, mark=square] file {ip/Russian.dat};
\addplot[smooth,color=violet, mark=triangle] file {ip/Spanish.dat};

\addplot[dashed, smooth,color=red,mark=oplus,mark options={scale=1,solid}] file {sp/English.dat};
%\addplot[dashed, smooth,color=blue,mark=otimes] file {ip/Chinese.dat};
\addplot[dashed, smooth,color=orange, fill opacity=0.1,mark=o,mark options={scale=1,solid}] file {sp/Arabic.dat};

\addplot[dashed, smooth,color=red!50!green,mark=x,mark options={scale=1,solid}] file {sp/Korean.dat};

\addplot[dashed, color=teal, fill opacity=0.1, mark=square,mark options={scale=1,solid}] file {sp/Russian.dat};
\addplot[dashed, smooth,color=violet, mark=triangle,mark options={scale=1,solid}] file {sp/Spanish.dat};

\addlegendentry{Arabic}
\addlegendentry{Chinese}
\addlegendentry{English}
\addlegendentry{Korean}
\addlegendentry{Russian}
\addlegendentry{Spanish}

\end{axis}
\end{tikzpicture}
}
\caption{Segmentation results with (dashed) and without space delimiters with different numbers of training instances $N$.}\label{fig:5b}
\end{figure}
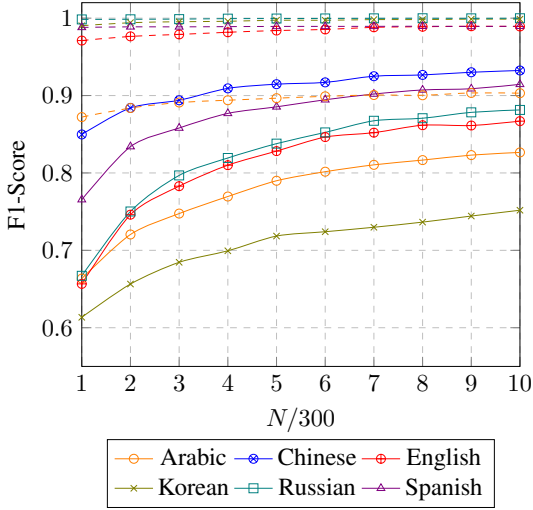

Chinese and Japanese are not delimited by spaces. Additionally, continuous writing without spaces (scriptio continua) is evidenced in most Classical Greek and Latin manuscripts. We perform two sets of learning curve experiments to investigate the impact of white space on word segmentation. In the first set, we keep the datasets in their original forms. In the second set, we omit all white space. The experimental results are presented in Figure \ref{fig:5b}. 

In general, there are huge discrepancies between the accuracies with and without spaces, showing that white space acts crucially as a word boundary indicator. Retaining the original forms of the space-delimited languages, very high accuracies can be achieved even with small amounts of training data as the model quickly learns that space is a reliable word boundary indicator. Moreover, we obtain relatively lower scores on space-delimited languages when space is ignored than Chinese using comparable amounts of training data, which shows that Chinese characters are more informative to word boundary prediction, due to the large character set  size. 

\subsubsection{Non-Segmental Multiword Tokens}\label{sec:mt}

\begin{figure}[t]
\centering
\scalebox{0.85}{
\begin{tikzpicture}
\begin{axis}[
%x=12cm,
xtick={1, 2,...,10},
ytick={0, 0.1, 0.2, 0.3, 0.4, 0.5, 0.6, 0.7, 0.8, 0.9, 1.0},  
ylabel=F1-Score,
xlabel=$N/300$, 
xmin=1,
xmax=10,
ymin=0.8,ymax=1.02,
ymajorgrids=true,
 xmajorgrids=true,
 grid style=dashed,
legend columns=3, 
%legend pos=outer north east,
%legend style={font=\fontsize{7}{6}}
legend style={at={(0.5, -0.2)},anchor=north}
]

\addplot[smooth,color=orange, fill opacity=0.1,mark=o] file {mwt/Arabic.dat};

\addplot[smooth,color=blue,mark=otimes] file {mwt/French.dat};
\addplot[smooth,color=red,mark=oplus] file {mwt/Hebrew.dat};
\addplot[smooth,color=red!50!green,mark=x] file {mwt/Italian.dat};
\addplot[smooth,color=teal, fill opacity=0.1, mark=square] file {mwt/Portuguese.dat};
\addplot[smooth,color=violet, mark=triangle] file {mwt/Spanish.dat};

\addplot[dashed,smooth,color=orange, fill opacity=0.1,mark=o,mark options={scale=1,solid}] file {iwt/Arabic.dat};

\addplot[dashed,smooth,color=blue,mark=otimes,mark options={scale=1,solid}] file {iwt/French.dat};
\addplot[dashed,smooth,color=red,mark=oplus,mark options={scale=1,solid}] file {iwt/Hebrew.dat};
\addplot[dashed,smooth,color=red!50!green,mark=x,mark options={scale=1,solid}] file {iwt/Italian.dat};
\addplot[dashed,smooth,color=teal, fill opacity=0.1, mark=square,mark options={scale=1,solid}] file {iwt/Portuguese.dat};
\addplot[dashed,smooth,color=violet, mark=triangle,mark options={scale=1,solid}] file {iwt/Spanish.dat};

\addlegendentry{Arabic}
\addlegendentry{French}
\addlegendentry{Hebrew}
\addlegendentry{Italian}
\addlegendentry{Portuguese}
\addlegendentry{Spanish}

\end{axis}
\end{tikzpicture}
}
\caption{Segmentation results with and without (dashed) processing non-segmental multiword tokens with different training instances $N$.}\label{fig:5c}
\end{figure}
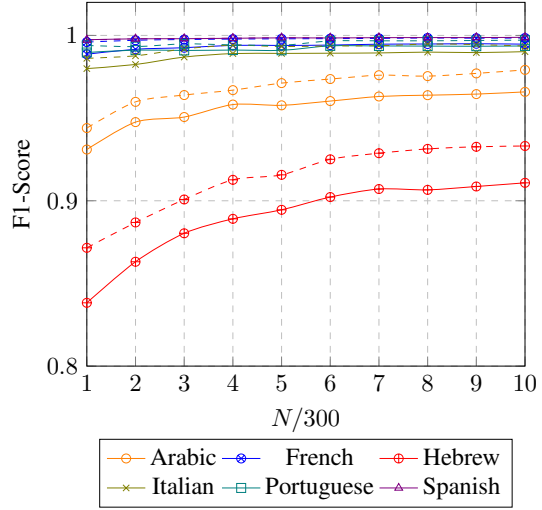

\begin{table}
\centering
\scalebox{0.85}{
\begin{tabular}{l|c|c|c|c}
\hline
\multirow{2}{*}{Language} & \multicolumn{2}{c|}{Data size} & \multicolumn{2}{c}{Evaluation Scores} \\
\cline{2-5}
 & Training & Validation & ACC & MFS \\
\hline
Arabic & 3,500 & 184 & 77.84 & 82.64  \\
\hline
Hebrew & 2,995 & 157 & 84.81 & 92.35\\
\hline
\end{tabular}
}
\caption{Accuracy of the seq2seq transducer on Arabic and Hebrew. }\label{tab:3}
\end{table}

\begin{table}[t]
\centering
\scalebox{0.88}{
\begin{tabular}{l|c|c|c|c}
\hline
 & None & Dictionary & Transducer & Mix \\
\hline
Arabic & 94.11 & 96.74 & 96.54 & \textbf{97.27}  \\
\hline
Hebrew & 87.17 & 91.33 & 88.46 & \textbf{91.85}\\
\hline
\end{tabular}
}
\caption{Segmentation accuracies on Arabic and Hebrew with different ways of transducing non-segmental multiword tokens.}\label{tab:3a}
\end{table}

The concept of multiword tokens is specific to UD. To explore how the non-segmental multiword tokens, as opposed to pure segmentation, influence segmentation accuracy, we conduct relevant experiments on selected languages. Similarly to the previous section, two sets of learning curve experiments are performed. In the second set, all the multiword tokens that require transduction are regarded as single words without being processed. The results are presented in Figure \ref{fig:5c}. 

Word segmentation with full UD processing is notably more challenging for Arabic and Hebrew. Table \ref{tab:3} shows the evaluation of the encoder-decoder as the transducer for non-segmental multiword tokens on Arabic and Hebrew. The evaluation metrics ACC and MF-score (MFS) are adapted from the metrics used for machine transliteration evaluation \cite{li2009report}. ACC is exact match and MFS is based on edit distance. The transducer yields relatively higher scores on Hebrew while it is more challenging to process Arabic.
%It is difficult to identify and transcribe the multiword tokens correctly. 

In addition, different approaches to transducing the non-segmental multiword tokens are evaluated in Table \ref{tab:3a}. In the condition None, the identified non-segmental multiword tokens remain unprocessed. In Dictionary, they are mapped via the dictionary derived from training data if found in the dictionary. In Transducer, they are all transduced by the attention-based encoder-decoder. In Mix, in addition to utilising the mapping dictionary, the non-segmental terms not found in the dictionary are transduced with the encoder-decoder. The results show that when the encoder-decoder is applied alone, it is worse than only using the dictionaries, but additional improvements can be obtained by combining both of them. 

The accuracy differences associated with non-segmental multiword tokens are nonetheless marginal on the other languages as shown in Figure \ref{fig:5c}. Regardless of their frequent occurrences, multiword tokens are easy to process in general when the set of unique non-segmental multiword tokens is small.

\subsubsection{Correlations with Accuracy}

\begin{figure}[t]
\centering

\begin{center}
\scalebox{0.90}{

\begin{tikzpicture}
\begin{axis}[
	ylabel,
	symbolic x coords={TS, CS, LS, AL, SF, MP, MS},
	enlargelimits=0.0,
	legend style={at={(0.5,-0.15)},
	anchor=north,legend columns=-1
	},
	ybar, 
	bar width=14pt, 
	enlarge x limits=0.1,
	ymin = -0.016,
	ymax = 0.016,
]
\addplot[fill=violet] 
coordinates {(TS, 0.00045635584728558634) (CS,  -0.001865200674309974)(LS, -0.00022728549775871538) (AL, 0.0003526089729863857) (SF,-0.015111926189820697) (MP, -0.001915767273039409) (MS, -0.00688313917856785)};
	%coordinates {(TS, 0.02550567) (CS,  0.01211569 )(LS, -0.00924826) (AL, -0.0024574) (SF,-0.02205263) (MP,  0.01184381) (MS, -0.00973657)};
\end{axis}
\end{tikzpicture}
}
\end{center}

\caption{Correlation coefficients between segmentation accuracy and the typological factors in the linear regression model. The factors are training set size (TS), character set size (CS), lexicon size (LS), average word length (AL), segmentation frequency (SF), multitoken word portion (MP) and multitoken word size (MS). }
\label{fig:5d2}
\end{figure}
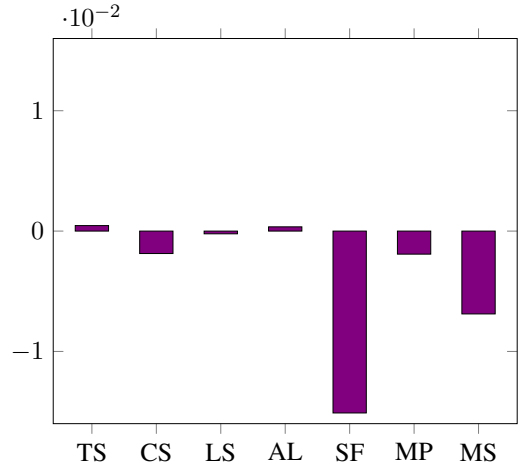

We investigate the correlations between the proposed typological factors in Section \ref{sec:1} and segmentation accuracy using linear regression with Huber loss \cite{huber1964robust}. The factors are used in addition to training set size as the features to predict the segmentation accuracies in F1-score. To collect more data samples, apart from experimenting with the full training data for each set, we also use smaller sets of 500, 1,000 and 2,000 training instances to train the models respectively if the training set is large enough. The features are standardised with the arithmetic mean and the standard deviation before fitting the linear regression model. 

The correlation coefficients of the linear regression model are presented in Figure \ref{fig:5d2}. We can see that segmentation frequency and multiword token set size are negatively correlated with segmentation accuracy.  Overall, the UD datasets are strongly biased towards space-delimited languages. Training set size is therefore not a strong factor as high accuracies can be obtained with small amounts of training data, which is consistent with the results of all the learning curve experiments. The other typological factors such as average word length and lexicon size are less relevant to segmentation accuracy. Referring back to Figure \ref{fig:3}, segmentation frequency and multiword token set size as the most influential factors, are also the primary principal components that categorise the UD languages into different groups. 

\subsubsection{Language-Specific Settings}

Our model obtains competitive results with only a minimal number of straightforward language-specific settings. Based on the previous analysis of segmentation accuracy and typological factors, referring back to Figure \ref{fig:3}, we apply the following settings, targeting on
specific language groups, to the segmentation system on the final test sets. The language-specific settings can be applied to new languages beyond the UD datasets based on an analysis of the typological factors.

\begin{enumerate}[noitemsep,topsep=0.1cm]
\item For languages with word-internal spaces like Vietnamese, we first separate punctuation and then use space-delimited syllables for boundary prediction. 
\item For languages with large character sets and no space delimiters, like Chinese and Japanese, we use concatenated 3-gram representations. 
\item For languages with more than 200 unique non-segmental multiword tokens, like Arabic and Hebrew, we use the encoder-decoder model for transduction.
\item For other languages, the universal model is sufficient without any specific adaptation. 
\end{enumerate}

\begin{table} 
\centering
\scalebox{0.87}{
\begin{tabular}{c|c|c|c}
\hline
Space & NLTK & UDPipe & This Paper \\
\hline
 80.86 & 95.64 & 99.47 & 99.45  \\
\hline
\end{tabular}
}
\caption{Average evaluation scores on UD languages, excluding Chinese, Japanese, Vietnamese, Arabic and Hebrew.} \label{tab:3j}
\end{table}

\begin{table*}[h]
\centering
\scalebox{0.72}{
\begin{tabular}{c|c|c|c|c|c|c|c|c}
\hline
Dataset & UDPipe & This Paper & Dataset & UDPipe & This Paper & Dataset & UDPipe & This Paper  \\
\hline
Ancient Greek & 99.98 & 99.96 & Ancient Greek-PROIEL & 99.99 & 100.0 & Arabic & \cellcolor{green! 43}93.77 & \cellcolor{green! 43}97.16 \\
Arabic-PUD & \cellcolor{green! 60}90.92 & \cellcolor{green! 60}95.93 & Basque & 99.97 & 100.0 & Bulgarian & 99.96 & 99.93 \\
Catalan & 99.98 & 99.80 & Chinese & \cellcolor{green! 43}90.47 & \cellcolor{green! 43}93.82 & Croatian & 99.88 & 99.95 \\
Czech & 99.94 & 99.97 & Czech-CAC & 99.96 & 99.93 & Czech-CLTT & 99.58 & 99.64 \\
Czech-PUD & 99.34 & 99.62 & Danish & 99.83 & 100.0 & Dutch & 99.84 & 99.92 \\
Dutch-LassySmall & 99.91 & 99.96 & English & 99.05 & 99.13 & English-LinES & 99.90 & 99.95 \\
English-PUD & 99.69 & 99.71 & English-ParTUT & 99.60 & 99.51 & Estonian & 99.90 & 99.88 \\
Finnish & 99.57 & 99.74 & Finnish-FTB & 99.95 & 99.99 & Finnish-PUD & 99.64 & 99.39 \\
French & \cellcolor{green! 15}98.81 & \cellcolor{green! 15}99.39 & French-PUD & \cellcolor{red! 26}98.84 & \cellcolor{red! 26}97.23 & French-ParTUT & \cellcolor{green! 13}98.97 & \cellcolor{green! 13}99.32 \\
French-Sequoia & \cellcolor{green! 13}99.11 & \cellcolor{green! 13}99.48 & Galician & 99.94 & 99.97 & Galician-TreeGal & \cellcolor{red! 15}98.66 & \cellcolor{red! 15}98.07 \\
German & 99.58 & 99.64 & German-PUD & 97.94 & 97.74 & Gothic & 100.0 & 100.0 \\
Greek & 99.94 & 99.86 & Hebrew & \cellcolor{green! 68}85.16 & \cellcolor{green! 68}91.01 & Hindi & 100.0 & 100.0 \\
Hindi-PUD & \cellcolor{green! 15}98.26 & \cellcolor{green! 15}98.82 & Hungarian & 99.79 & 99.93 & Indonesian & 100.0 & 100.0 \\
Irish & \cellcolor{green! 14}99.38 & \cellcolor{green! 14}99.85 & Italian & 99.83 & 99.54 & Italian-PUD & \cellcolor{red! 14}99.21 & \cellcolor{red! 14}98.78 \\
Japanese & \cellcolor{green! 27}92.03 & \cellcolor{green! 27}93.77 & Japanese-PUD & \cellcolor{green! 15}93.67 & \cellcolor{green! 15}94.17 & Kazakh & 94.17 & 94.21 \\
Korean & 99.73 & 99.95 & Latin & 99.99 & 100.0 & Latin-ITTB & 99.94 & 100.0 \\
Latin-PROIEL & 99.90 & 100.0 & Latvian & \cellcolor{green! 13}99.16 & \cellcolor{green! 13}99.56 & Norwegian-Bokmaal & 99.83 & 99.89 \\
Norwegian-Nynorsk & 99.91 & 99.97 & Old Church Slavonic & 100.0 & 100.0 & Persian & 99.65 & 99.62 \\
Polish & 99.90 & 99.93 & Portuguese & \cellcolor{red! 14}99.59 & \cellcolor{red! 14}99.10 & Portuguese-BR & \cellcolor{red! 13}99.85 & \cellcolor{red! 13}99.52 \\
Portuguese-PUD & \cellcolor{red! 14}99.40 & \cellcolor{red! 14}98.98 & Romanian & 99.68 & 99.74 & Russian & 99.66 & 99.96 \\
Russian-PUD & 97.09 & 97.28 & Russian-SynTagRus & 99.64 & 99.65 & Slovak & 100.0 & 99.98 \\
Slovenian & 99.93 & 100.0 & Slovenian-SST & 99.91 & 100.0 & Spanish & 99.75 & 99.85 \\
Spanish-AnCora & 99.94 & 99.93 & Spanish-PUD & 99.44 & 99.39 & Swedish & 99.79 & 99.97 \\
Swedish-LinES & 99.93 & 99.98 & Swedish-PUD & \cellcolor{green! 19}98.36 & \cellcolor{green! 19}99.26 & Turkish & 98.09 & 97.85 \\
Turkish-PUD & \cellcolor{red! 13}96.99 & \cellcolor{red! 13}96.68 & Ukrainian & 99.81 & 99.76 & Urdu & 100.0 & 100.0 \\
\cline{7-9}
Uyghur & \cellcolor{red! 29}99.85 & \cellcolor{red! 29}97.86 & Vietnamese & \cellcolor{green! 32}85.53 & \cellcolor{green! 32}87.79 & \textbf{Average} & 98.63 & \textbf{98.90} \\
\hline
\end{tabular}

}
\caption{Evaluation results on the UD test sets in F1-scores. The datasets are represented in the corresponding treebank codes. \textit{PUD} suffix indicates the parallel test data. Two shades of green/red are used for better visualisation, with brighter colours for larger differences. Green represents that our system is better than UDPipe and red is used otherwise.}\label{tab:4}
\end{table*}

\begin{table}[t]
\centering
\scalebox{0.87}{
\begin{tabular}{l|c|c|c|c}
\hline
 & BT & BT+CRF & FT & FT+CRF\\
 \hline
Chinese & 90.54 & 90.66 & 90.73 & \textbf{91.28}  \\
\hline
Japanese & 91.54 & 91.64 & 91.88 & \textbf{91.94}\\
\hline
Vietnamese & 87.63 & \textbf{87.95} & 87.61 & 87.75  \\
\hline
Arabic & 94.47 & 96.74 & 94.73 & \textbf{97.16}  \\
\hline
Hebrew & 85.34 & 90.74 & 85.53 & \textbf{91.98}\\
\hline
\end{tabular}
}
\caption{Comparison between the binary tags (BT) and the fine-grained tags (FT) as well as the effectiveness of the CRF interface on the development sets.}\label{tab:3d}
\end{table}

\begin{table}[t]
\centering
\scalebox{0.87}{
\begin{tabular}{l|c|c|c|c}
\hline
 & Arabic & French & German & Hebrew\\
 \hline
UDPipe & 79.34 & \textbf{98.91} & 94.21 & 71.87  \\
\hline
Our model & \textbf{91.35} & 97.50 & 94.21 & \textbf{86.17}  \\
\hline
\end{tabular}
}
\caption{Percentages of the correctly processed multiword tokens on the development sets.}\label{tab:3f}
\end{table}

\subsection{Final Results}
\label{sec:63}

\begin{table*} 
\centering
\scalebox{0.78}{
\begin{tabular}{l|cc|cc|cc|cc|cc}
\hline
& \multicolumn{2}{c|}{Segmentation} & \multicolumn{4}{c|}{UDPipe parser} & \multicolumn{4}{c}{\newcite{dozat-qi-manning:2017:K17-3}} \\
\cline{4-11}
& \multicolumn{2}{c|}{Accuracy} & \multicolumn{2}{c|}{UAS} & \multicolumn{2}{c|}{LAS} & \multicolumn{2}{c|}{UAS} & \multicolumn{2}{c}{LAS} \\
\hline
 & UDPipe & This Paper & UDPipe & This Paper & UDPipe & This Paper & UDPipe & This Paper & UDPipe & This Paper\\
\hline
Arabic & 93.77 & 97.16 & 72.34 & 78.22 & 66.41 & 71.79 & 77.52 & 83.55 &  72.89 & 78.42 \\
Chinese & 90.47 & 93.82 & 63.20 & 67.91 & 59.07 & 63.31 & 71.24 & 76.33 & 68.20  & 73.04 \\
Hebrew & 85.16 & 91.01 & 62.14 & 71.18 & 57.82 & 66.59 & 67.61  & 76.39 & 64.02  & 72.37  \\
Japanese & 92.03 & 93.77 & 78.08 & 81.77 &  76.73 & 80.83 & 80.21 &  83.79 & 79.44 &  82.99 \\
Vietnamese & 85.53 & 87.79 & 47.72 & 50.87 & 43.10 & 46.03 & 50.28 & 53.78 &  45.54  & 48.86 \\
\hline
\end{tabular}
}
\caption{Extrinsic evaluations with dependency parsing on the test sets. The parsing accuracies are reported in unlabelled attachment score (UAS) and labelled attachment score (LAS).} \label{tab:3i}
\end{table*}

We compare our segmentation model to UDPipe \cite{udpipe:2017} on the test sets. UDPipe contains word segmentation, POS tagging, morphological analysis and dependency parsing models in a pipeline.
 The word segmentation model in UDPipe is also based on RNN with GRU. For efficiency, UDPipe has a smaller character embedding size and no CRF interface. It also relies heavily on white-space and uses specific configurations for languages in which word-internal spaces are allowed. Automatically generated suffix rules are applied jointly with a dictionary query to handle multiword tokens. Moreover, UDPipe uses language-specific hyper-parameters for Chinese and Japanese.

%We employ the pretained UDPipe models on UD 2.0 that are publicly available. 
%We employ the publicly available UDPipe models trained on UD v2.0. 
We employ UDPipe 1.2 with the publicly available UD 2.0 models.\footnote{http://hdl.handle.net/11234/1-2364}
The \textit{presegmented} option is enabled as we assume the input text to be presegmented into sentences so that only word segmentation is evaluated. In addition, the CoNLL shared task involved some test sets for which no specific training data were available. This included a number of parallel test sets of known languages, for which we apply the models trained on the standard treebanks, as well as four surprise languages, namely Buryat, Kurmanji, North Sami and Upper Sorbian, for which we use the small annotated data samples provided in addition to the test sets by the shared task to build models and evaluation on those languages.

%for our system, we also use the small annotated data samples of the four surprise languages, namely Buryat, Kurmanji, North Sami and Upper Sorbian that are provided by the CoNLL 2017 shared task \cite{zeman-EtAl:2017:K17-3} as the training sets to build models and evaluate on these languages. For both UDPipe and our segmenter, we apply the models trained on the standard treebanks to the parallel test data. 

The main evaluation results are shown in Table \ref{tab:4}. We also report the Macro Average F1-scores. The scores of the surprise languages are excluded and presented separately as no corresponding UDPipe models are available. 

Our system obtains higher segmentation accuracy overall. It achieves substantially better accuracies on languages that are challenging to segment, namely Chinese, Japanese, Vietnamese, Arabic and Hebrew. The two systems yield very similar scores, when these languages are excluded as shown in Table \ref{tab:3j}, in which the two systems are also compared with two rule-based baselines, a simple space-based tokeniser and the tokenisation model for English in NLTK \cite{loper2002nltk}. The NLTK model obtains relatively high accuracy while the space-based baseline substantially underperforms, which indicates that relying on white space alone is insufficient for word segmentation in general. On the majority of the space-delimited languages without productive non-segmental multiword tokens, both UDPipe and our segmentation system yield near-perfect scores in Table \ref{tab:4}. In general, referring back to Figure \ref{fig:3}, languages that are clustered at the bottom-left corner are relatively trivial to segment. 

The evaluation scores are notably lower on Semitic languages as well as languages without word delimiters. Nonetheless, our system obtains substantially higher scores on the languages that are more challenging to process. 

For Chinese, Japanese and Vietnamese, our system benefits substantially from the concatenated 3-gram character representation, which has been demonstrated in Section \ref{sec:ng}. Besides, we employ a more fine-grained tagset with CRF loss instead of the binary tags adopted in UDPipe. As presented in \newcite{Y06-1012}, more fine-grained tagging schemes outperform binary tags, which is supported by the experimental results on morpheme segmentation reported in \newcite{ruokolainen2013supervised}. 

We further investigate the merits of the fine-grained tags over the binary tags as well as the effectiveness of the CRF interface by the experiments presented in Table \ref{tab:3d} with the variances of our segmentation system. The fine-grained tags denote the boundary tags introduced in Table \ref{tab:1}. The binary tags include two basic tags {\tt B}, {\tt I} plus the corresponding tags {\tt \textoverline{B}}, {\tt  \textoverline{I}} for non-segmental multiword tokens. White space is marked as {\tt I} instead of {\tt X}. The concatenated 3-grams are not applied. In general, the experimental results confirm that the fine-grained tags are more beneficial except for Vietnamese. The fine-grained tagset contains more structured positional information that can be exploited by the word segmentation model. Additionally, the CRF interface leads to notable improvements, especially for Arabic and Hebrew. The combination of the fine-grained tags with the CRF interface achieves substantial improvements over the basic binary tag model that is analogous to UDPipe.  
%We can see that the structured information provided by the fine-grained tags is helpful in general and can be effectively utilised by the CRF interface, whereas adding CRF does not lead to notable improvements when the binary tags are used. The combination of the fine-grained tags with the CRF interface achieves substantial improvements over the basic binary tag model that is analogous to UDPipe,  not only for Chinese, Japanese and Vietnamese, but also for Arabic and Hebrew. 

For Arabic and Hebrew, apart from greatly benefiting from the fine-grained tagset and the CRF interface, our model is better at handling non-segmental multiword tokens (Table \ref{tab:3f}). 
%UDPipe is not capable of disambiguating the multiword tokens that should be processed differently with respect to the context. 
The attention-based encoder-decoder as the transducer is much more powerful in processing the non-segmental multiword tokens that are not covered by the dictionary than the suffix rules for analysing multiword tokens in UDPipe.  

UDPipe obtains higher scores on a few datasets. Our model overfits the small training data of Uyghur as it yields 100.0 F1-score on the development set. For a few parallel test sets, there are punctuation marks not found in the training data that cannot be correctly analysed by our system as it is fully data-driven without any heuristic rules for unknown characters.

\begin{table} 
\centering
\scalebox{0.75}{
\begin{tabular}{l|c|c|c|l}
\hline
& Space & NLTK & Sample & Transfer \\
\hline
Buryat & 71.99 & 97.99 & 88.07 & 97.99 (Russian) \\
Kurmanji & 78.97 & \textbf{97.37} & 93.37 & 96.71 (Spanish)\\
North Sami & 79.07 & 99.20 & 92.82 & \textbf{99.81} (German) \\
Upper Sorbian & 72.35 & \textbf{94.60} & 93.34 & 93.66 (Spanish)\\
\hline
\end{tabular}
}
\caption{Evaluation on the surprise languages.} \label{tab:3g}
\end{table}

%bxr & - & 88.07 & hsb & - & 93.34 & kmr & - & 93.37 \\
%sme & - & 92.82 &  & & & & & \\

The evaluation results on the surprise languages are presented in Table \ref{tab:3g}. In addition to the segmentation models proposed in this paper, we present the evaluation scores of a space-based tokeniser as well as the NLTK model for English. As shown by the previous learning curve experiments in Section \ref{sec:52}, very high accuracies can be obtained on the space-delimited languages with only small amounts of training data. However, in case of extreme data sparseness (less than 20 training sentences), such as for the four surprise languages in Table \ref{tab:3g} and Kazakh in Table \ref{tab:4}, the segmentation results are drastically lower even though the surprise languages are all space-delimited. 
%However, in the extreme cases as the four surprise languages as well as Kazakh in Table \ref{tab:4}, in which less than 20 sentences are used for training, the segmentation results are drastically lower than the other languages with sufficient training data even though they are all space-delimited.  

For the surprise languages, we find that applying segmentation models trained on a different language with more training data yields better results than relying on the small annotated samples of the target language.
%we also transfer the segmentation models trained on languages with large annotated data apart from utilising the small annotated samples for training specific models. 
Considering that the segmentation model is fully character-based, we simply select the model of the language that shares the most characters with the surprise language as its segmentation model. No annotated data of the surprise language are used for model selection. As shown in Table \ref{tab:3g}, the transfer approach achieves comparable segmentation accuracies to NLTK. For space-delimited languages with insufficient training data, it may be beneficial to employ a well-designed rule-based word segmenter as NLTK occasionally outperforms the data-driven approach.
%, indicating that process any space-delimited language with the existing models even if no annotated data are available, which underlines the universal character of the proposed segmentation system.

%In addition, we evaluate the segmenter with dependency parsing as the subsequent task on the datasets where we obtained substantial improvements over UDPipe. Apart from the parsing model in UDPipe 1.2, we employ the graph-based parser by \newcite{dozat-qi-manning:2017:K17-3} for evaluation. The experimental results are shown in Table \ref{tab:3i}. We can see that word segmentation accuracy has a great impact on parsing accuracy as the segmentation errors propagate. Having a more accurate word segmentation model is very beneficial for achieving higher parsing accuracy. 

As a form of extrinsic evaluation, we test the segmenter in a dependency parsing setup on the datasets where we obtained substantial improvements over UDPipe. We present results for the transition-based parsing model in UDPipe 1.2 and for the graph-based parser by \newcite{dozat-qi-manning:2017:K17-3}. The experimental results are shown in Table \ref{tab:3i}. We can see that word segmentation accuracy has a great impact on parsing accuracy as the segmentation errors propagate. Having a more accurate word segmentation model is very beneficial for achieving higher parsing accuracy.

\section{Related Work}

The BiRNN-CRF model is proposed by \newcite{huang2015bidirectional} and has been applied to a number of sequence labelling tasks, such as part-of-speech tagging, chunking and named entity recognition.

Our universal word segmenter is a major extension of the joint word segmentation and POS tagging system described by \newcite{shao17}. The original model is specifically developed for Chinese and only applicable to Chinese and Japanese. Apart from being language-independent, the proposed model in this paper employs an extended tagset and a complementary sequence transduction component to fully process non-segmental multiword tokens that are present in a substantial amount of languages, such as Arabic and Hebrew in particular. It is a generalised segmentation and transduction framework.

Our universal model is compared with the language-specific model of \newcite{shao17} in Table \ref{tab:3h}.  With pretrained character embeddings, ensemble decoding and joint POS tags prediction as introduced in \newcite{shao17}, considerable improvements over the universal model presented in this paper can be obtained. However, the joint POS tagging system is difficult to generalise as single characters in space-delimited languages are usually not informative for POS tagging. Additionally, compared to Chinese, sentences in space-delimited languages have a much greater number of characters on average. Combining the POS tags with segmentation tags drastically enlarges the search space and therefore the model becomes extremely inefficient both for training and tagging. The joint POS tagging model is nonetheless applicable to Japanese and Vietnamese.

\begin{table} 
\centering
\scalebox{0.86}{
\begin{tabular}{l|c|c|c|c}
\hline
& This Paper & Shao  & Che  & Bj\"{o}rkelund  \\
\hline
Chinese & 93.82 & 95.21 & 91.19 & 92.81  \\
Japanese & 93.77 & 94.79 & 92.95 & 91.68 \\
\hline
Arabic & 97.16 & -- & 93.71 & 95.53 \\
Hebrew & 91.01 & -- & 85.16 & 91.37 \\
\hline
\end{tabular}
}
\caption{Comparison between the universal model and the language-specific models. } \label{tab:3h}
\end{table}

\newcite{monroe2014word} present a data-driven word segmentation system for Arabic based on a sequence labelling framework. An extended tagset is designed for Arabic-specific orthographic rules and applied together with hand-crafted features in a CRF framework. It obtains 98.23 F1-score on newswire Arabic Treebank,\footnote{LDC2010T13, LDC2011T09, LDC2010T08} 97.61 on Broadcast News Treebank,\footnote{LDC2012T07} and 92.10 on the Egyptian Arabic dataset.\footnote{LDC2012E{93,98,89,99,107,125}, LDC2013E{12,21}} For Hebrew, \newcite{Goldberg:2013:WSU:2464100.2464107} perform word segmentation jointly with syntactic disambiguation using lattice parsing. Each lattice arc corresponds to a word and its corresponding POS tag, and a path through the lattice corresponds to a specific word segmentation and POS tagging of the sentence. The proposed model is evaluated on the Hebrew Treebank \cite{Guthmann20081A}. The joint word segmentation and parsing F1-score (76.95) is reported and compared against the parsing score (85.70) with gold word segmentation. The evaluation scores reported in both \newcite{monroe2014word} and \newcite{Goldberg:2013:WSU:2464100.2464107} are not directly comparable to the evaluation scores on Arabic and Hebrew in this paper, as they are obtained on different datasets.

For universal word segmentation, apart from UDPipe described in Section~\ref{sec:63}, there are several systems that are developed for specific language groups. \newcite{che-EtAl:2017:K17-3} build a similar Bi-LSTM word segmentation model targeting languages without space delimiters like Chinese and Japanese. The proposed model incorporates rich statistics-based features gathered from large-scale unlabelled data, such as character unigram embeddings, character bigram embeddings and the point-wise mutual information of adjacent characters. \newcite{bjorkelund-EtAl:2017:K17-3} use a CRF-based tagger for multiword token rich languages like Arabic and Hebrew. A predicted Levenshtein edit script is employed to transform the multiword tokens into their components. The evaluation scores on a selected set of languages reported in \newcite{che-EtAl:2017:K17-3} and \newcite{bjorkelund-EtAl:2017:K17-3} are included in Table \ref{tab:3h} as well. 

 \newcite{more2018} adapt existing morphological analysers for Arabic, Hebrew and Turkish and present ambiguous word segmentation possibilities for these languages in a lattice format (CoNLL-UL) that is compatible with UD. The CoNLL-UL datasets can be applied as external resources for processing non-segmental multiword tokens.\footnote{CoNLL-UL is not evaluated in our experiments as it is very recent work.} 

%In addition, the BiRNN-CRF model as the fundamental segmentation framework has been applied to a number of sequence labelling tasks, such as part-of-speech tagging, chunking and named entity recognition .

\section{Conclusion}

We propose a sequence tagging model and apply it to universal word segmentation. BiRNN-CRF is adopted as the fundamental segmentation framework that is complemented by an attention-based sequence-to-sequence transducer for non-segmental multiword tokens. We propose six typological factors to characterise the difficulty of word segmentation cross different languages. The experimental results show that segmentation accuracy is primarily correlated with segmentation frequency as well as the set of non-segmental multiword tokens. Using whitespace as delimiters is crucial to word segmentation, even if the correlation between orthographic tokens and words is not perfect. For space-delimited languages, very high accuracy can be obtained even with relatively small training sets, while more training data is required for high segmentation accuracy for languages without spaces. Based on the analysis, we apply a minimal number of language-specific settings to substantially improve the segmentation accuracy for languages that are fundamentally more difficult to process.

The segmenter is extensively evaluated on the UD datasets in various languages and compared with UDPipe. Apart from obtaining nearly perfect segmentation on most of the space-delimited languages, our system achieves high accuracies on languages without space delimiters such as Chinese and Japanese as well as Semitic languages with abundant multiword tokens like Arabic and Hebrew.

%With respect to the experimental results on the correlations between typology and segmentation accuracies, white spaces as word delimiters are crucial to word segmentation. For space-delimited languages, very high accuracy can be obtained even with relatively small training set, while more training data are required for high segmentation accuracy if we artificially omit white spaces in their original forms. Languages with higher segmentation frequency are notably more challenging to segment in general. Additionally, characters in languages with larger character sets are more informative to word segmentation. Moreover, as a UD specific definition, the size of multiword token set is negatively related to segmentation accuracy very substantially, whereas the portion is less relevant. 

\section*{Acknowledgments}

We acknowledge the computational resources provided by CSC in Helsinki and Sigma2 in Oslo through NeIC-NLPL (www.nlpl.eu). This work is supported by the Chinese Scholarship Council (CSC) (No. 201407930015). We would like to thank the TACL editors and reviewers for their valuable feedback.

\bibliography{tacl}

\begin{thebibliography}{}

\bibitem[\protect\citename{Abadi \bgroup et al.\egroup
  }2016]{abadi2016tensorflow}
Mart{\'\i}n Abadi, Paul Barham, Jianmin Chen, Zhifeng Chen, Andy Davis, Jeffrey
  Dean, Matthieu Devin, Sanjay Ghemawat, Geoffrey Irving, Michael Isard, et~al.
\newblock 2016.
\newblock Tensor{F}low: A system for large-scale machine learning.
\newblock In {\em Proceedings of the 12th USENIX Symposium on Operating Systems
  Design and Implementation (OSDI)}, pages 265--283.

\bibitem[\protect\citename{Abdi and Williams}2010]{abdi2010principal}
Herv{\'e} Abdi and Lynne~J Williams.
\newblock 2010.
\newblock Principal component analysis.
\newblock {\em Wiley Interdisciplinary Reviews: Computational Statistics},
  2(4):433--459.

\bibitem[\protect\citename{Bahdanau \bgroup et al.\egroup
  }2015]{bahdanau2014neural}
Dzmitry Bahdanau, Kyunghyun Cho, and Yoshua Bengio.
\newblock 2015.
\newblock Neural machine translation by jointly learning to align and
  translate.
\newblock In {\em International Conference on Learning Representations}.

\bibitem[\protect\citename{Bj\"{o}rkelund \bgroup et al.\egroup
  }2017]{bjorkelund-EtAl:2017:K17-3}
Anders Bj\"{o}rkelund, Agnieszka Falenska, Xiang Yu, and Jonas Kuhn.
\newblock 2017.
\newblock {IMS} at the {C}o{NLL} 2017 {UD} shared task: {CRF}s and perceptrons
  meet neural networks.
\newblock In {\em Proceedings of the CoNLL 2017 Shared Task: Multilingual
  Parsing from Raw Text to Universal Dependencies}, pages 40--51.

\bibitem[\protect\citename{Che \bgroup et al.\egroup
  }2017]{che-EtAl:2017:K17-3}
Wanxiang Che, Jiang Guo, Yuxuan Wang, Bo~Zheng, Huaipeng Zhao, Yang Liu,
  Dechuan Teng, and Ting Liu.
\newblock 2017.
\newblock The {HIT-SCIR} system for end-to-end parsing of universal
  dependencies.
\newblock In {\em Proceedings of the {C}o{NLL} 2017 Shared Task: Multilingual
  Parsing from Raw Text to {U}niversal {D}ependencies}, pages 52--62.

\bibitem[\protect\citename{Chen \bgroup et al.\egroup }2015]{chen2015long}
Xinchi Chen, Xipeng Qiu, Chenxi Zhu, Pengfei Liu, and Xuanjing Huang.
\newblock 2015.
\newblock Long short-term memory neural networks for {C}hinese word
  segmentation.
\newblock In {\em Conference on Empirical Methods in Natural Language
  Processing}, pages 1197--1206.

\bibitem[\protect\citename{Cho \bgroup et al.\egroup }2014]{cho2014properties}
Kyunghyun Cho, Bart Van~Merri{\"e}nboer, Dzmitry Bahdanau, and Yoshua Bengio.
\newblock 2014.
\newblock On the properties of neural machine translation: Encoder-decoder
  approaches.
\newblock {\em arXiv preprint arXiv:1409.1259}.

\bibitem[\protect\citename{de Lhoneux \bgroup et al.\egroup }2017]{uu-conll17}
Miryam de~Lhoneux, Yan Shao, Ali Basirat, Eliyahu Kiperwasser, Sara Stymne,
  Yoav Goldberg, and Joakim Nivre.
\newblock 2017.
\newblock From raw text to {U}niversal {D}ependencies -- look, no tags!
\newblock In {\em Proceedings of the CoNLL 2017 Shared Task: Multilingual
  Parsing from Raw Text to Universal Dependencies.}, pages 207--217.

\bibitem[\protect\citename{Dozat \bgroup et al.\egroup
  }2017]{dozat-qi-manning:2017:K17-3}
Timothy Dozat, Peng Qi, and Christopher~D. Manning.
\newblock 2017.
\newblock Stanford's graph-based neural dependency parser at the {C}o{NLL} 2017
  shared task.
\newblock In {\em Proceedings of the CoNLL 2017 Shared Task: Multilingual
  Parsing from Raw Text to Universal Dependencies}, pages 20--30, Vancouver,
  Canada, August. Association for Computational Linguistics.

\bibitem[\protect\citename{Duchi \bgroup et al.\egroup
  }2011]{duchi2011adaptive}
John Duchi, Elad Hazan, and Yoram Singer.
\newblock 2011.
\newblock Adaptive subgradient methods for online learning and stochastic
  optimization.
\newblock {\em Journal of Machine Learning Research}, 12(Jul):2121--2159.

\bibitem[\protect\citename{Glorot and Bengio}2010]{glorot2010understanding}
Xavier Glorot and Yoshua Bengio.
\newblock 2010.
\newblock Understanding the difficulty of training deep feedforward neural
  networks.
\newblock In {\em International Conference on Artificial Intelligence and
  Statistics}, pages 249--256.

\bibitem[\protect\citename{Goldberg and
  Elhadad}2013]{Goldberg:2013:WSU:2464100.2464107}
Yoav Goldberg and Michael Elhadad.
\newblock 2013.
\newblock Word segmentation, unknown-word resolution, and morphological
  agreement in a {H}ebrew parsing system.
\newblock {\em Computational Linguistics}, 39(1):121--160, March.

\bibitem[\protect\citename{Guthmann \bgroup et al.\egroup
  }2009]{Guthmann20081A}
Noemie Guthmann, Yuval Krymolowski, Adi Milea, and Yoad Winter.
\newblock 2009.
\newblock Automatic annotation of morphosyntactic dependencies in a modern
  {H}ebrew.
\newblock In {\em Proceedings of the 1st Workshop on Treebanks and Linguistic
  Theories}, pages 1--12.

\bibitem[\protect\citename{Hochreiter and Schmidhuber}1997]{hochreiter1997long}
Sepp Hochreiter and J{\"u}rgen Schmidhuber.
\newblock 1997.
\newblock Long short-term memory.
\newblock {\em Neural computation}, 9(8):1735--1780.

\bibitem[\protect\citename{Huang \bgroup et al.\egroup
  }2015]{huang2015bidirectional}
Zhiheng Huang, Wei Xu, and Kai Yu.
\newblock 2015.
\newblock Bidirectional {LSTM-CRF} models for sequence tagging.
\newblock {\em arXiv preprint arXiv:1508.01991}.

\bibitem[\protect\citename{Huber}1964]{huber1964robust}
Peter~J Huber.
\newblock 1964.
\newblock Robust estimation of a location parameter.
\newblock {\em The annals of mathematical statistics}, pages 73--101.

\bibitem[\protect\citename{Itai and Wintner}2008]{itai2008language}
Alon Itai and Shuly Wintner.
\newblock 2008.
\newblock Language resources for {H}ebrew.
\newblock {\em Language Resources and Evaluation}, 42(1):75--98.

\bibitem[\protect\citename{Lafferty \bgroup et al.\egroup
  }2001]{lafferty2001crf}
John~D. Lafferty, Andrew McCallum, and Fernando C.~N. Pereira.
\newblock 2001.
\newblock Conditional random fields: Probabilistic models for segmenting and
  labeling sequence data.
\newblock In {\em Proceedings of the Eighteenth International Conference on
  Machine Learning}, ICML '01, pages 282--289.

\bibitem[\protect\citename{Li \bgroup et al.\egroup }2009]{li2009report}
Haizhou Li, A~Kumaran, Vladimir Pervouchine, and Min Zhang.
\newblock 2009.
\newblock Report of {NEWS} 2009 machine transliteration shared task.
\newblock In {\em Proceedings of the 2009 Named Entities Workshop: Shared Task
  on Transliteration}, pages 1--18.

\bibitem[\protect\citename{Loper and Bird}2002]{loper2002nltk}
Edward Loper and Steven Bird.
\newblock 2002.
\newblock {NLTK}: The natural language toolkit.
\newblock In {\em Proceedings of the ACL-02 Workshop on Effective tools and
  methodologies for teaching natural language processing and computational
  linguistics-Volume 1}, pages 63--70. Association for Computational
  Linguistics.

\bibitem[\protect\citename{Monroe \bgroup et al.\egroup }2014]{monroe2014word}
Will Monroe, Spence Green, and Christopher~D Manning.
\newblock 2014.
\newblock Word segmentation of informal arabic with domain adaptation.
\newblock In {\em Proceedings of the 52nd Annual Meeting of the Association for
  Computational Linguistics (Volume 2: Short Papers)}, volume~2, pages
  206--211.

\bibitem[\protect\citename{More \bgroup et al.\egroup }2018]{more2018}
Amir More, \"{O}zlem \c{C}etino\u{g}lu, \c{C}a\u{g}r{\i} \c{C}\"{o}ltekin,
  Nizar Habash, Beno\^{i}t Sagot, Djam\'{e} Seddah, Dima Taji, and Reut
  Tsarfaty.
\newblock 2018.
\newblock {C}o{NLL-UL}: Universal morphological lattices for {U}niversal
  {D}ependency parsing.
\newblock In {\em Proceedings of the Eleventh International Conference on
  Language Resources and Evaluation}.

\bibitem[\protect\citename{Nivre \bgroup et al.\egroup
  }2016]{nivre2016universal}
Joakim Nivre, Marie-Catherine de~Marneffe, Filip Ginter, Yoav Goldberg, Jan
  Hajic, Christopher~D. Manning, Ryan McDonald, Slav Petrov, Sampo Pyysalo,
  Natalia Silveira, Reut Tsarfaty, and Daniel Zeman.
\newblock 2016.
\newblock Universal dependencies v1: A multilingual treebank collection.
\newblock In {\em Proceedings of the 10th International Conference on Language
  Resources and Evaluation}, pages 1659--1666.

\bibitem[\protect\citename{Palmer and Burger}1997]{palmer1997chinese}
David Palmer and John Burger.
\newblock 1997.
\newblock Chinese word segmentation and information retrieval.
\newblock In {\em AAAI Spring Symposium on Cross-Language Text and Speech
  Retrieval}, pages 175--178.

\bibitem[\protect\citename{Ruokolainen \bgroup et al.\egroup
  }2013]{ruokolainen2013supervised}
Teemu Ruokolainen, Oskar Kohonen, Sami Virpioja, and Mikko Kurimo.
\newblock 2013.
\newblock Supervised morphological segmentation in a low-resource learning
  setting using conditional random fields.
\newblock In {\em Proceedings of the Seventeenth Conference on Computational
  Natural Language Learning}, pages 29--37, Sofia, Bulgaria. Association for
  Computational Linguistics.

\bibitem[\protect\citename{Sag \bgroup et al.\egroup }2002]{sag2002multiword}
Ivan~A Sag, Timothy Baldwin, Francis Bond, Ann Copestake, and Dan Flickinger.
\newblock 2002.
\newblock Multiword expressions: A pain in the neck for {NLP}.
\newblock In {\em International Conference on Intelligent Text Processing and
  Computational Linguistics}, pages 1--15. Springer.

\bibitem[\protect\citename{Shao \bgroup et al.\egroup }2017]{shao17}
Yan Shao, Christian Hardmeier, J{\"o}rg Tiedemann, and Joakim Nivre.
\newblock 2017.
\newblock Character-based joint segmentation and {POS} tagging for {C}hinese
  using bidirectional {RNN-CRF}.
\newblock In {\em Proceedings the 8th International Joint Conference on Natural
  Language Processing}, pages 173--183.

\bibitem[\protect\citename{Srivastava \bgroup et al.\egroup
  }2014]{srivastava2014dropout}
Nitish Srivastava, Geoffrey~E Hinton, Alex Krizhevsky, Ilya Sutskever, and
  Ruslan Salakhutdinov.
\newblock 2014.
\newblock Dropout: a simple way to prevent neural networks from overfitting.
\newblock {\em Journal of Machine Learning Research}, 15(1):1929--1958.

\bibitem[\protect\citename{Straka and Strakov\'{a}}2017]{udpipe:2017}
Milan Straka and Jana Strakov\'{a}.
\newblock 2017.
\newblock Tokenizing, {POS} tagging, lemmatizing and parsing {UD} 2.0 with
  {UDP}ipe.
\newblock In {\em Proceedings of the CoNLL 2017 Shared Task: Multilingual
  Parsing from Raw Text to Universal Dependencies}, pages 88--99.

\bibitem[\protect\citename{Xue}2003]{xue2003chinese}
Nianwen Xue.
\newblock 2003.
\newblock Chinese word segmentation as character tagging.
\newblock {\em Computational Linguistics and Chinese Language Processing},
  pages 29--48.

\bibitem[\protect\citename{Zeman \bgroup et al.\egroup
  }2017]{zeman-EtAl:2017:K17-3}
Daniel Zeman, Martin Popel, Milan Straka, Jan Hajic, Joakim Nivre, Filip
  Ginter, Juhani Luotolahti, Sampo Pyysalo, Slav Petrov, Martin Potthast,
  Francis Tyers, Elena Badmaeva, Memduh Gokirmak, Anna Nedoluzhko, Silvie
  Cinkova, Jan Hajic~jr., Jaroslava Hlavacova, V\'{a}clava Kettnerov\'{a},
  Zdenka Uresova, Jenna Kanerva, Stina Ojala, Anna Missil\"{a}, Christopher~D.
  Manning, Sebastian Schuster, Siva Reddy, Dima Taji, Nizar Habash, Herman
  Leung, Marie-Catherine de~Marneffe, Manuela Sanguinetti, Maria Simi, Hiroshi
  Kanayama, Valeria dePaiva, Kira Droganova, H\'{e}ctor Mart\'{i}nez~Alonso,
  \c{C}a\u{g}rı \c{C}\"{o}ltekin, Umut Sulubacak, Hans Uszkoreit, Vivien
  Macketanz, Aljoscha Burchardt, Kim Harris, Katrin Marheinecke, Georg Rehm,
  Tolga Kayadelen, Mohammed Attia, Ali Elkahky, Zhuoran Yu, Emily Pitler, Saran
  Lertpradit, Michael Mandl, Jesse Kirchner, Hector~Fernandez Alcalde, Jana
  Strnadov\'{a}, Esha Banerjee, Ruli Manurung, Antonio Stella, Atsuko Shimada,
  Sookyoung Kwak, Gustavo Mendonca, Tatiana Lando, Rattima Nitisaroj, and Josie
  Li.
\newblock 2017.
\newblock Co{NLL} 2017 shared task: Multilingual parsing from raw text to
  {U}niversal {D}ependencies.
\newblock In {\em Proceedings of the CoNLL 2017 Shared Task: Multilingual
  Parsing from Raw Text to Universal Dependencies}, pages 1--19.

\bibitem[\protect\citename{Zhao \bgroup et al.\egroup }2006]{Y06-1012}
Hai Zhao, Chang-Ning Huang, Mu~Li, and Bao-Liang Lu.
\newblock 2006.
\newblock Effective tag set selection in {C}hinese word segmentation via
  conditional random field modeling.
\newblock In {\em Proceedings of the 20th Pacific Asia Conference on Language,
  Information and Computation}, pages 87--94.

\bibitem[\protect\citename{Zhou}1991]{zhou1991}
Youguang Zhou.
\newblock 1991.
\newblock The family of {C}hinese character-type scripts.
\newblock {\em Sino-Platonic Papers, 28}.

\end{thebibliography}
\bibliographystyle{acl2012}

%\newpage

%\ 
%\vspace{5mm}

%\newpage

%\onecolumn
%\section*{Appendix}

%\begin{table*}[!th]

%\begin{center}

%\scalebox{0.70}{
%\input{stats.tex}
%}
%\caption{Statistics on the training sets of UD 2.0 regarding training set size (TS), character set size (CS), lexicon size (LS), average word length (AL), segmentation frequency (SF), multiword token portion (MP) and multiword token size (MS). The intra-language differences are mainly caused by the genre variation of the treebanks.}\label{tab:7}
%\end{center}
%\end{table*}

\end{CJK}

\end{document}